\documentclass[10pt,twocolumn,letterpaper]{article}

\usepackage{iccv}              

\definecolor{cvprblue}{rgb}{0.21,0.49,0.74}
\usepackage[pagebackref,breaklinks,colorlinks,citecolor=cvprblue]{hyperref}

\usepackage{amsfonts}       
\usepackage{nicefrac}       
\usepackage{microtype}      
\usepackage{xcolor}         
\usepackage{multirow}
\usepackage{graphicx}
\usepackage{arydshln}

\usepackage{wrapfig}
\usepackage{lipsum} 
\usepackage{booktabs} 
\usepackage{algorithm}
\usepackage{algpseudocode}

\usepackage{listings}

\definecolor{best}{HTML}{fdf4bb}
\definecolor{gemini}{HTML}{bffdbb}

\definecolor{tfblue}{RGB}{21,101,192}
\definecolor{tfgreen}{RGB}{56,142,60}
\definecolor{tfyellow}{RGB}{255,193,7}
\definecolor{tfred}{RGB}{233,30,99}
\definecolor{tfpurple}{RGB}{103,58,183}
\definecolor{tfgray}{RGB}{97,97,97}

\lstdefinestyle{tfstyle}{
    language=Python,
    basicstyle=\ttfamily\scriptsize,
    keywordstyle=\color{tfblue},
    stringstyle=\color{tfgreen},
    commentstyle=\color{tfgray},
    morecomment=[l][\color{tfgray}]{\#},
    morekeywords={tf, import, class, def},
    morestring=[b]",
    backgroundcolor=\color{white},
    frame=tb,
    showstringspaces=false,
    breaklines=true,
    captionpos=b,
    numbers=left,
    numberstyle=\tiny\color{tfgray},
    emph={self},
    emphstyle=\color{tfred},
}

\def\paperID{644} 
\def\confName{ICCV}
\def\confYear{2025}

\title{What Do You See?\\Enhancing Zero-Shot Image Classification with Multimodal Large Language Models}

\author{Abdelrahman Abdelhamed\thanks{Equal contribution.}\\
Google Research\\
{\tt\small aabdelhamed@google.com}
\and
Mahmoud Afifi\footnotemark[1]\\
Google\\
{\tt\small mafifi@google.com}
\and
Alec Go\\
Google Research\\
{\tt\small ago@google.com}
}

\begin{document}
\maketitle

\begin{abstract}
Large language models (LLMs) have been effectively used for many computer vision tasks, including image classification. In this paper, we present a simple yet effective approach for zero-shot image classification using multimodal LLMs. Using multimodal LLMs, we generate comprehensive textual representations from input images. These textual representations are then utilized to generate fixed-dimensional features in a cross-modal embedding space. Subsequently, these features are fused together to perform zero-shot classification using a linear classifier. Our method does not require prompt engineering for each dataset; instead, we use a single, straightforward set of prompts across all datasets. We evaluated our method on several datasets and our results demonstrate its remarkable effectiveness, surpassing benchmark accuracy on multiple datasets. On average, for ten benchmarks, our method achieved an accuracy gain of 6.2 percentage points, with an increase of 6.8 percentage points on the ImageNet dataset, compared to prior methods re-evaluated with the same setup. Our findings highlight the potential of multimodal LLMs to enhance computer vision tasks such as zero-shot image classification, offering a significant improvement over traditional methods.
\end{abstract}

\section{Introduction and related work}
\label{sec:intro}

\begin{figure}[t]
\centering
\includegraphics[width=\linewidth]{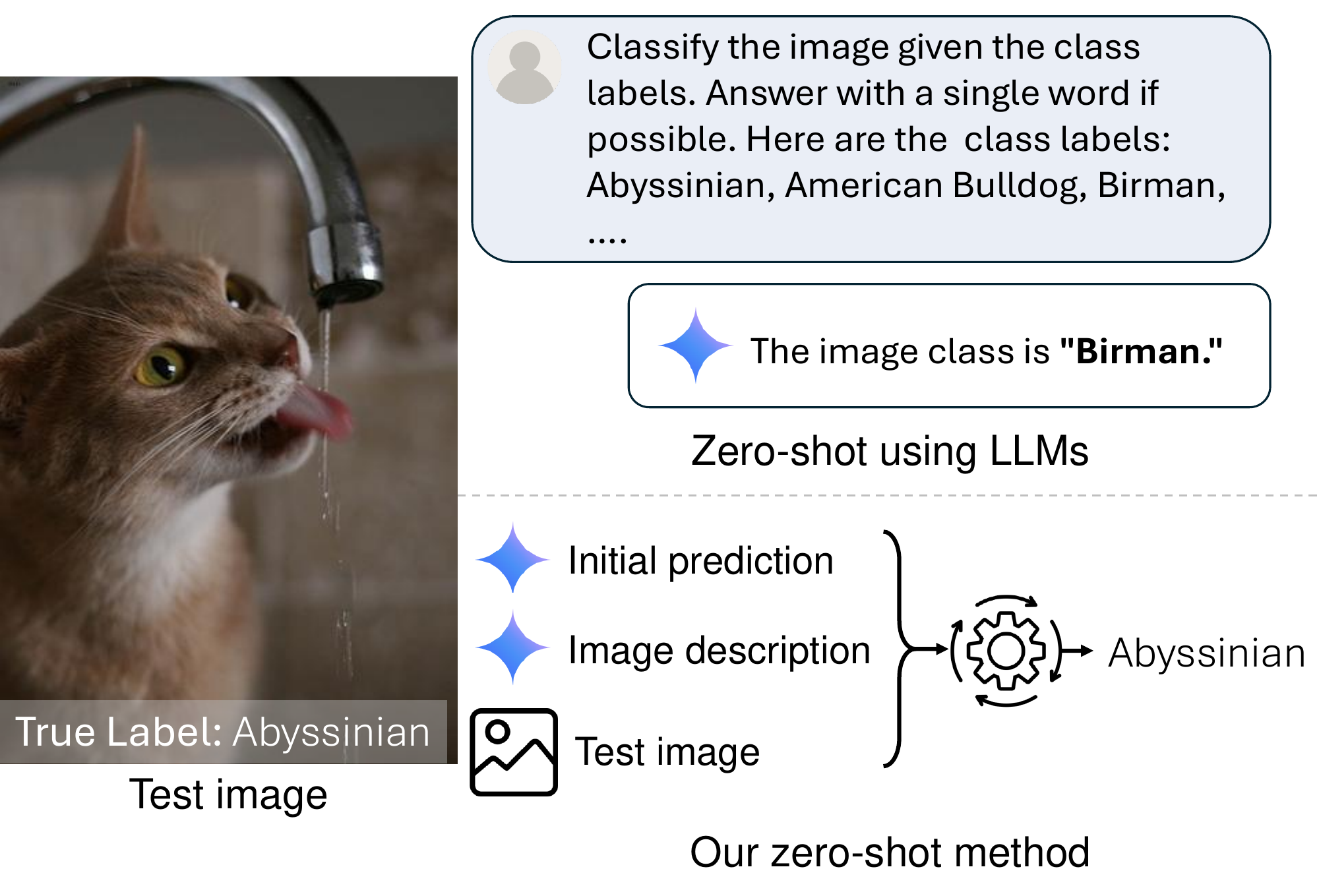}
\vspace{-3mm}
\caption{\label{fig:intro}While LLMs can perform zero-shot image classification, they may not consistently generate accurate target dataset labels or follow instructions to produce only the class name. Our method incorporates LLMs by leveraging initial predictions from such models, along with image descriptions and image data, to produce accurate and consistent class labels. Shown image is sourced from the Pets dataset \cite{pets}.}
\end{figure}

Zero-shot image classification aims to categorize images into classes unseen during training, presenting a significant challenge in computer vision. Recent approaches leverage the power of large language models (LLMs) like GPT-4 \cite{GPT} to generate prompts for target classes, often in conjunction with vision-language models (VLMs) such as CLIP \cite{CLIP} to embed images and text in a common space. Open-vocabulary models like CLIP \cite{CLIP} and VirTex \cite{VirTex} have shown promise in this area due to their ability to generalize to unseen classes. These models learn to match images with captions from vast amounts of image-text data, allowing for dynamic classification without retraining. Early works like DeViSE \cite{DeViSE} pioneered the concept of a joint embedding space for images and text, enabling generalization to unseen classes. Approaches such as CLIP \cite{CLIP}, based on contrastive learning, and ALIGN \cite{ALIGN}, employing a two-stage framework, further refined the alignment of image and text representations. More recent approaches for zero-shot image classification (e.g., \cite{CuPL, menon2022visual}) have utilized LLMs to generate prompts (i.e., captions or descriptions) for the target classes to further improve the classification accuracy.

Using only the visual content of images to make predictions can reduce accuracy because images alone often do not contain enough information to match the detailed meanings found in text descriptions. 
To address this limitation, we propose our method to leverage the capabilities of \textit{multimodal} LLMs to generate rich textual representations of the input images. Multimodal LLMs, such as GPT-4 \cite{GPT} and Gemini \cite{gemini}, have demonstrated remarkable abilities in various tasks.  They are capable of processing and integrating information from various sources such as text, images, and audio. This allows them to perform tasks that were previously challenging, such as generating detailed image descriptions, answering complex visual questions, and even creating realistic images from text.   Inspired by these advancements, we utilize a straightforward set of prompts to generate detailed textual descriptions of the input images (see Figure~\ref{fig:intro}), eliminating the need for complex prompt engineering seen in previous works (e.g., \cite{CLIP, CALIP}). These textual representations are then fused with visual features to perform zero-shot classification.  

While zero-shot image classification using large language models (LLMs) initially appeared promising, our experiments demonstrate insufficient accuracy.  Even with careful prompt engineering, LLMs frequently predict classes outside the defined target set. Therefore, we propose a refined approach: encoding LLM predictions as input features for a linear classifier. This classifier constrains the output to the target class set, improving prediction accuracy. Our method offers several key advantages:
\begin{itemize}
    \item \textit{Improved accuracy} by leveraging rich textual information extracted directly from images, significantly boosting classification accuracy.

    \item \textit{Constrained LLM predictions} to restrict LLM predictions to the desired target classes.

    \item Simplified prompting using a set of universal prompts, eliminating the need for dataset-specific prompts.

    \item Superior performance in a consistent experimental setup, outperforming existing methods across various benchmark datasets. Specifically, our method achieves an average accuracy gain of 6.2 percentage points in ten image classification benchmark datasets and an accuracy increase of 6.8\% in the ImageNet dataset \cite{imagenet}.
\end{itemize}

In the following sections, we detail our proposed approach for zero-shot image classification using LLMs (Section \ref{sec:method}), present experimental results across ten benchmark datasets and analyze the computational resources used (Section \ref{sec:experiments}), discuss limitations (Section \ref{sec:limitations}), and conclude with remarks and future directions (Section \ref{sec:conclusion}).

\section{Method}
\label{sec:method}

\begin{figure*}[!t]
\centering
\includegraphics[width=\linewidth]{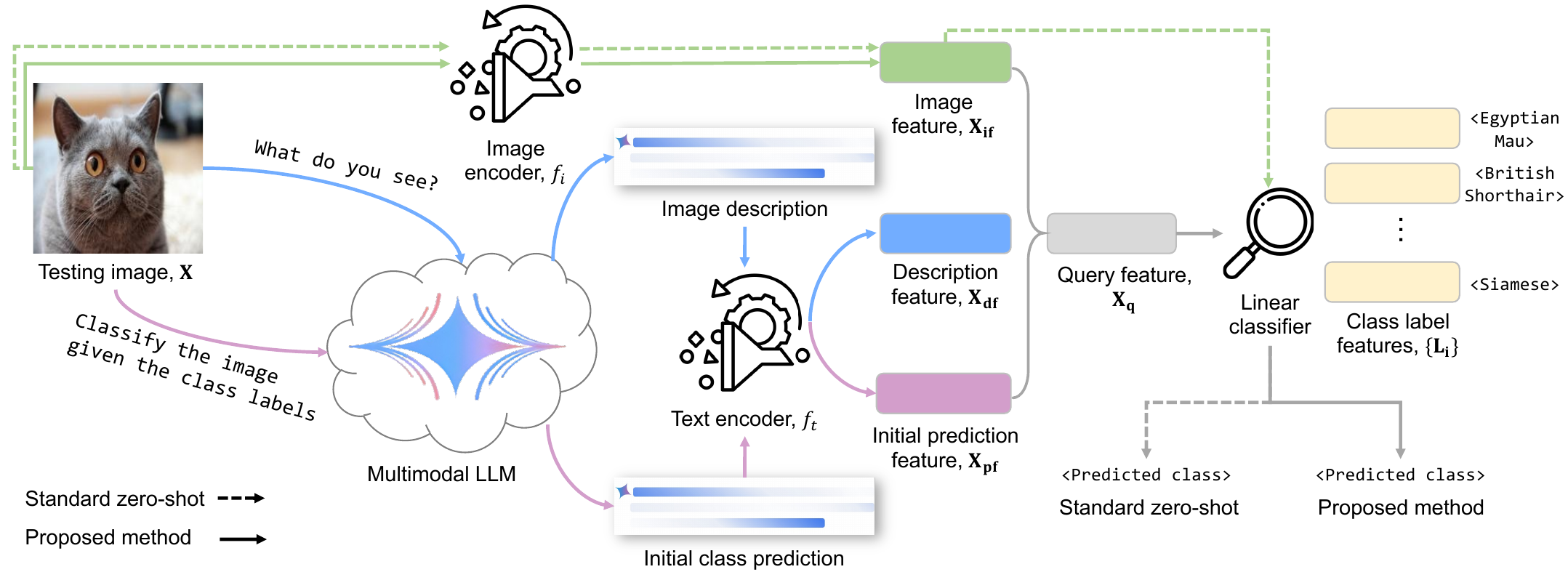}
\vspace{-3mm}
\caption{We propose a zero-shot image classification method that leverages multimodal large language models (LLMs) to enhance the accuracy of standard zero-shot classification. Our method employs a set of engineered prompts to generate image description and initial class prediction by the LLM. Subsequently, we encode this data along with the input testing image using a cross-modal embedding encoder to project the inputs into a common feature space. Finally, we fuse the generated features to produce the final query feature, which is then utilized by a standard zero-shot linear image classifier to predict the final class. The shown image is sourced from the Pets dataset \cite{pets}.}
\vspace{-3mm}
\label{fig:teaser}
\end{figure*}

Given an input image, $\textbf{X}$, containing object(s) belonging to a single class label from a finite set of class labels, $\{l_i\}_{i=1}^{m}$, our objective is to classify $\textbf{X}$ without any dataset-specific training process for image classification. The overview of our method is illustrated in Figure \ref{fig:teaser}. As shown in Figure \ref{fig:teaser}, our approach relies on a cross-modal embedding encoder models (image encoder, $f_i$, and text encoder, $f_t$), trained to learn joint representations of images and text, as demonstrated in prior works, such as \cite{CLIP, DeViSE, ALIGN}. Additionally, we utilize a multimodal LLM, $g$, which is pre-trained on a large corpus of multimodal data. This model, $g$, is designed to generate responses that align with both textual and visual inputs, effectively integrating information from both modalities to make predictions (e.g., Gemini \cite{gemini}).

\subsection{Class label features}

Our zero-shot classifier includes class label features of the target dataset. We first encode the set of class labels, $\{l_i\}_{i=1}^{m}$, in some embedding space by the cross-modal text encoder model, $f_t$, such that $\{\textbf{L}_i\}_{i=1}^{m}$ refers to the set of encoded class label features, where $\textbf{L}_i \in \mathbb{R}^n$ is a normalized $n$-D class label feature of the class label $l_i$ and $n$ is the dimensionality of the embedded features. 

Our zero-shot classifier uses the embedded class labels by representing $\{\textbf{L}_i\}$ as a 2D matrix, $\textbf{M} \in \mathbb{R}^{n \times m}$, by stacking all encoded class label features:

\begin{equation}
    \label{eq:if0}
    \textbf{M} = [\textbf{L}_1, \dots, \textbf{L}_m].
\end{equation}

Such an encoded feature matrix can be generated using one of three options: (1) directly from the textual class labels, (2) using a human-designed template (e.g., ``{\texttt{A photo of \{class\_label\}}}'' \cite{CLIP, CALIP}), where {\texttt{\{class\_label}\}} refers to the textual label of one of our classes $\{l_i\}_{i=1}^{m}$, or (3) LLM-generated class description(s) \cite{CuPL}. In option (3), the LLM-generated class description(s) are then converted to embedded features, followed by fusion (e.g., averaging) to generate a single embedded feature for each class label in the dataset.
Optionally, all features from the three options can be fused together (e.g., averaged) for increased robustness.

\subsection{Cross-modal input features}

\begin{figure*}[!t]
\centering
\includegraphics[width=\linewidth]{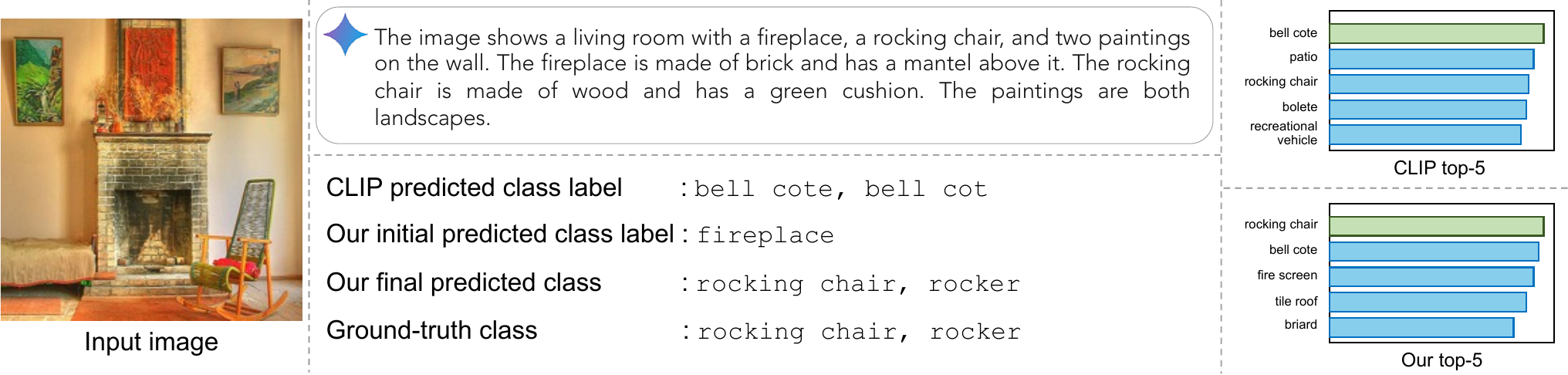}
\caption{Our method utilizes image description and initial class prediction generated by LLM, in addition to the input image, to improve the zero-shot classification accuracy of cross-modal embedding models, such as CLIP \cite{CLIP}. The shown image is from the ImageNet dataset \cite{imagenet}.}
\label{fig:results}
\end{figure*}

To predict the final class, we first encode the input image by the cross-modal image encoder model, $f_i$, to generate the image feature $\textbf{X}_{\texttt{if}} \in \mathbb{R}^n$ as follows:

\begin{equation}
    \label{eq:if1}
    \widetilde{\textbf{X}}_{\texttt{if}} = f_i\left(\textbf{X}\right),
\end{equation}

\begin{equation}
    \label{eq:if2}
    \textbf{X}_{\texttt{if}} =  \frac{1}{\|\widetilde{\textbf{X}}_{\texttt{if}}\|} \widetilde{\textbf{X}}_{\texttt{if}},
\end{equation}

where $\|\cdot\|$ performs vector normalization. Traditionally, the image feature serves as the sole input to prior zero-shot image classifiers \cite{CuPL, CLIP, CALIP}. Our method enhances this input by incorporating the LLM, $g$, which generates additional textual-based inputs for our zero-shot image classifier (see Figure \ref{fig:results}). To achieve this, we employ an engineered prompt that instructs the LLM to describe the input image, $\textbf{X}$, and perform initial image classification using the textual names of the class label set $\{l_i\}_{i=1}^{m}$. Denoting $p_{\texttt{d}}$ and $p_{\texttt{c}}$ as our prompts for image description and initial image classification, respectively, we generate two additional embedded features alongside $\textbf{X}_{\texttt{if}}$ as follows:

\begin{equation}
    \label{eq:df1}
    \widetilde{\textbf{X}}_{\texttt{df}} = \left(f_t\circ g\right)\left(\textbf{X}, p_{\texttt{d}}\right),
\end{equation}

\begin{equation}
    \label{eq:df2}
    \textbf{X}_{\texttt{df}} =  \frac{1}{\|\widetilde{\textbf{X}}_{\texttt{df}}\|} \widetilde{\textbf{X}}_{\texttt{df}},
\end{equation}

\begin{equation}
    \label{eq:pf1}
    \widetilde{\textbf{X}}_{\texttt{pf}} = \left(f_t\circ g\right)\left(\textbf{X}, p_{\texttt{c}}\right),
\end{equation}

\begin{equation}
    \label{eq:pf2}
    \textbf{X}_{\texttt{pf}} =  \frac{1}{\|\widetilde{\textbf{X}}_{\texttt{pf}}\|} \widetilde{\textbf{X}}_{\texttt{pf}},
\end{equation}

\noindent where $\textbf{X}_{\texttt{df}} \in \mathbb{R}^n$ and $\textbf{X}_{\texttt{pf}} \in \mathbb{R}^n$ refer to the image description feature and initial class prediction feature, respectively. Notably, in our method, unlike prior methods (e.g., \cite{CuPL, CLIP, CALIP}), such prompts $p_{\texttt{d}}$ and $p_{\texttt{c}}$  do not require dataset-specific engineering for each dataset. Instead, we employ fixed prompts: the first prompt instructs the LLM to provide a generic image description, while the second prompt includes the textual class labels of the target dataset.

\begin{table}[t]
\centering
\caption{\label{tab:prompts}Details of prompts utilized in our work. Each row represents one query task to the LLM. For instance,  `image classification' indicates the utilization of LLM to conduct initial zero-shot image classification, which serves as one of the features in our method. Both image classification and description can be combined in one query. The {\ttfamily{\{classes\}}} variable refers to the class labels of the dataset. The {\ttfamily{\{predicted\_class\}}} refers to Gemini's output of the image classification prompt. The {\ttfamily{\{class\_label\}}} variable denotes one of the class labels in the given dataset.}
\small
\begin{tabular}{p{.9\linewidth}}
  \toprule
Image classification \\\hline
  \ttfamily{You are given an image and a list of class labels. Classify the image given the class labels. Answer using a single word if possible. Here are the class labels: \{classes\}} \\\toprule 
Image description \\\hline
  \ttfamily{What do you see? Describe any object precisely, including its type or class.} \\\toprule 
Class description  \\\hline
1. \ttfamily{Describe what a \{class\_label\} looks like in one or two sentences.} \\
2. \ttfamily{How can you identify a \{class\_label\} in one or two sentences?} \\
3. \ttfamily{What does a \{class\_label\} look like? Respond with one or two sentences.} \\
4. \ttfamily{Describe an image from the internet of a \{class\_label\}. Respond with one or two sentences.} \\
5. \ttfamily{A short caption of an image of a \{class\_label\}:} 
\end{tabular}
\end{table}

\begin{figure*}[!t]
\centering
\includegraphics[width=\linewidth]{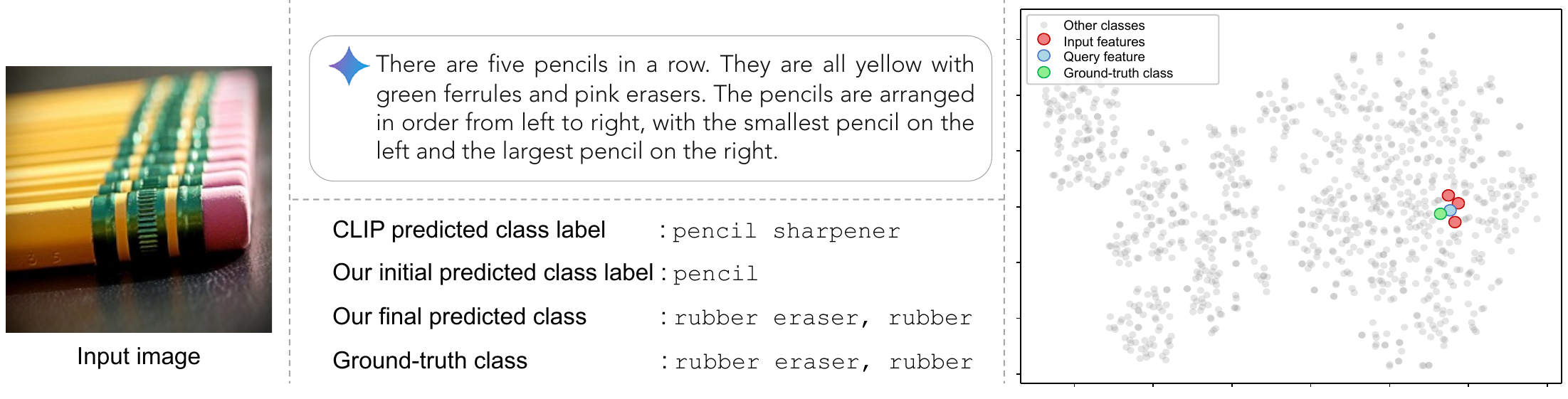}
\caption{Our query feature is a fusion of features extracted from the input image, image description, and initial prediction. This fusion operates similarly to ensembling, where our fused query feature demonstrates better robustness, achieving higher accuracy compared to traditional image features used in cross-modal-based zero-shot image classification (e.g., CLIP \cite{CLIP}). On the right, the t-SNE plot \cite{van2008visualizing} shows the class-embedded features of the ImageNet dataset \cite{imagenet} (in gray, with the ground-truth class of the shown image in green), our input features (in red), and the query feature after fusion (in blue).}
\label{fig:fusion}
\end{figure*}

After generating the three input features (image feature, description feature, and initial prediction feature), we fuse them to generate our final query feature, $\textbf{X}_{\texttt{q}}$. One can interpret this fusion as an ensemble of different candidate features to generate a more precise query feature (see Figure \ref{fig:fusion}). We adopted a simple averaging fusion, where the final query feature, $\textbf{X}_{\texttt{q}}$, is generated by:

\begin{equation}
\label{eq:fusion1}
\widetilde{\textbf{X}}_{\texttt{q}} = {\textbf{X}_{\texttt{if}} + \textbf{X}_{\texttt{df}} + \textbf{X}_{\texttt{pf}}} \ ,
\end{equation}

\begin{equation}
    \label{eq:fusion2}
    \textbf{X}_{\texttt{q}} =  \frac{1}{\|\widetilde{\textbf{X}}_{\texttt{q}}\|} \widetilde{\textbf{X}}_{\texttt{q}} \ .
\end{equation}

We found that this simple fusion yields good results compared to alternative fusion approaches. Refer to Section \ref{sec:ablations} for ablation studies.

\subsection{Class label prediction}

After computing our query feature, $\textbf{X}_{\texttt{q}}$, we apply our zero-shot linear classifier weights, $\textbf{M}$, to the fused query feature, $\textbf{X}_{\texttt{q}}$, to generate the final similarity scores (i.e., ``logits'') of our prediction. This process can be described as follows:

\begin{equation}
\label{eq:prediction}
\textbf{W} = \textbf{X}_{\texttt{q}}^T \textbf{M},
\end{equation}

\noindent where $T$ represents the vector transpose operation to transform $\textbf{X}_{\texttt{q}}$ into a row vector of shape $1 \times n$, and $\textbf{W} \in \mathbb{R}^{1\times m}$ contains the similarity scores of the generated query feature to the class label features in the target dataset. The index of the final predicted class is then computed as $\texttt{argmax}\left(\textbf{W}\right)$ that corresponds to the maximum similarity score. The inference process of our method is concisely described in Algorithm~\ref{alg:method}.

\begin{algorithm}[t]
  \caption{\label{alg:method}Performs zero-shot image classification.}
  \begin{algorithmic}
    \Require {Image $\textbf{X}$, class labels $\{l_i\}_{i=1}^{m}$, class label feature matrix $\textbf{M}$, multimodal LLM $g$, cross-modal encoders $f_i$ \& $f_t$, initial class prediction prompt $p_{\texttt{c}}$, image description prompt $p_{\texttt{d}}$ }
    
    $\widetilde{\textbf{X}}_{\texttt{if}} = f_i\left(\textbf{X}\right)$ \Comment{Image feature}
    
    $\textbf{X}_{\texttt{if}} =  \widetilde{\textbf{X}}_{\texttt{if}} / \|\widetilde{\textbf{X}}_{\texttt{if}}\|  $ \Comment{Vector normalization}
    
    $\widetilde{\textbf{X}}_{\texttt{df}} = \left(f_t\circ g\right)\left(\textbf{X}, p_{\texttt{d}}\right)$  \Comment{Image description feature}
    
    $\textbf{X}_{\texttt{df}} =  \widetilde{\textbf{X}}_{\texttt{df}} / \|\widetilde{\textbf{X}}_{\texttt{df}}\|  $ \Comment{Vector normalization}
    
    $\widetilde{\textbf{X}}_{\texttt{pf}} = \left(f_t\circ g\right)\left(\textbf{X}, p_{\texttt{c}}\right)$ \Comment{Initial prediction feature}
    
    $\textbf{X}_{\texttt{pf}} =  \widetilde{\textbf{X}}_{\texttt{pf}} / \|\widetilde{\textbf{X}}_{\texttt{pf}}\|  $ \Comment{Vector normalization}
    
    $\widetilde{\textbf{X}}_{\texttt{q}} = {\textbf{X}_{\texttt{if}} + \textbf{X}_{\texttt{df}} + \textbf{X}_{\texttt{pf}}}$  \Comment{Fused feature}
    
    $\textbf{X}_{\texttt{q}} =  \widetilde{\textbf{X}}_{\texttt{q}} / \|\widetilde{\textbf{X}}_{\texttt{q}}\|  $ \Comment{Vector normalization}
    
    $\textbf{W} = \textbf{X}_{\texttt{q}}^T \textbf{M}$ \Comment{Similarity scores}
    
    $x \leftarrow \texttt{argmax}\left(\textbf{W}\right)$ \Comment{Predicted class index}
    
    \Ensure {Predicted class label $l_x$ of input image}
  \end{algorithmic}
\end{algorithm}

\section{Experiments}
\label{sec:experiments}

In our experiments, we employed Gemini \cite{gemini} as our multimodal LLM, $g$, for generating image descriptions and initial predictions. Results using GPT \cite{GPT} are provided in the supplemental materials. We utilized CLIP (ViT-L/14) \cite{CLIP} as our cross-modal embedding encoder models, $f_i$ and $f_t$, to encode input testing images, image descriptions, and initial predictions generated by Gemini.

We explored four different versions of class label features, $\textbf{M}$ (Equation \ref{eq:if0}). Specifically, we used CLIP to encode the following representations of class labels: 1) class label names, 2) the text template {\texttt{``A photo of \{class\_label\}''}}, where {\texttt{\{class\_label\}}} denotes each class label in each dataset, and 3) class descriptions, similar to those generated by CuPL \cite{CuPL}, and 4) a combination of the aforementioned three options---akin to the fusion of our input features, we combined the three encoded features of each class and computed their average feature. The class descriptions were produced by prompting Gemini to describe each class label in the dataset 50 times, resulting in 50 different descriptions for each class. Subsequently, we utilized CLIP to encode the 50 class descriptions for each class and compute the average encoded feature vector to represent each class label. The class description features are generated once as described in Section \ref{sec:method}. 

Table \ref{tab:prompts} shows the prompts used for each step in our method employing a multimodal LLM (Gemini \cite{gemini}). The table shows the prompts used to: (1) perform zero-shot image classification with Gemini \cite{gemini}, (2) describe a given testing image, and (3) generate class label descriptions. Both image classification and description can be combined in one prompt. The class descriptions were generated using five prompts, with 10 responses generated for each prompt and class, resulting in 50 class descriptions per class label. To encourage diversity in Gemini's responses, we set the temperature parameter to a high value of 0.99, as done in \cite{CuPL}. 

\subsection{Results}
\label{sec:results}
We evaluated our method on the following datasets: ImageNet \cite{imagenet}, Pets \cite{pets}, Places365 \cite{places}, Food-101 \cite{food}, SUN397 \cite{sun1, sun2}, Stanford Cars \cite{cars}, Describable Textures Dataset (DTD) \cite{dtd}, Caltech-101 \cite{caltech101}, CIFAR-10 \cite{cifar}, and CIFAR-100 \cite{cifar}. We compared our results against two baselines using the same backbone as our method (ViT-L/14): (1) CLIP \cite{CLIP} and (2) CuPL \cite{CuPL}. The rationale for selecting these baselines is as follows. First, CLIP is a core component of our method, and comparing against it highlights the impact of our additional features and design choices. Second, CuPL extends CLIP by LLM to enrich the feature representations of target class labels. Since our method also employs an LLM for zero-shot classification, CuPL serves as a strong comparison baseline.

For CLIP \cite{CLIP}, we used a fixed text template, identical to our method, to ensure a fair comparison. Specifically, we employed the template ``{\texttt{A photo of \{class\_label\}}}'' to encode textual class labels across all datasets, except for Pets \cite{pets}, DTD \cite{dtd}, and Cars \cite{cars} datasets. For these exceptions, we employed specific templates: ``\texttt{A photo of \{class\_label\}, a type of pets}'', ``{\texttt{A photo of \{class\_label\}, a textural category}}'', and ``{\texttt{A photo of \{class\_label\}, a car model}}'', respectively. This approach was found to enhance results, consistent with prior findings \cite{CLIP, li2023your, allingham2023simple, popp2024zero}. For CuPL \cite{CuPL}, we replaced GPT-3 \cite{GPT} with Gemini \cite{gemini} to generate class descriptions, ensuring a fair comparison with our method, which also utilizes Gemini \cite{gemini}.

Additionally, it is worth noting that the results reported in the original CuPL paper \cite{CuPL} were computed using modified class labels for the ImageNet dataset \cite{imagenet}, differing from the dataset’s original labels. To ensure consistency and fairness, we re-evaluated the baseline methods using the standard ImageNet class labels while following the experimental setup described above. Additionally, for uniformity, we resized all images to 224$\times$224 before processing them with our method, CLIP \cite{CLIP}, and CuPL \cite{CuPL}.

\begin{table*}[t]
\centering
\caption{\label{tab:results}Comparison of classification accuracy between our method and baseline methods (CLIP \cite{CLIP} and CuPL \cite{CuPL}) across various datasets, including ImageNet \cite{imagenet}, CIFAR-10 (C-10) \cite{cifar}, CIFAR-100 (C-100) \cite{cifar}, Food-101 \cite{food}, SUN397 \cite{sun1, sun2}, Cars \cite{cars}, DTD \cite{dtd}, Caltech-101 \cite{caltech101}, Pets \cite{pets}, and Places \cite{places}. We report our results with the following class label features: 1) class descriptions, 2) class labels, 3) the template ``{\texttt{A photo of \{class\}}}'', and combined features of (1-3). The best results are highlighted in \colorbox{best}{\textbf{yellow}}.}
\vspace{-2mm}
\scalebox{0.8}{
\begin{tabular}{l|cccccccccc}
 &
  \multicolumn{9}{c}{\textbf{Dataset} (\textcolor{red}{number of classes}/\textcolor{blue}{number of testing images})} \\ \hline
\textbf{Method} &
 \multicolumn{1}{l|}{\begin{tabular}[c]{@{}c@{}}ImageNet\\(\textcolor{red}{\small{1K}}/\textcolor{blue}{\small{50K}})\end{tabular}} &
  \multicolumn{1}{l|}{\begin{tabular}[c]{@{}c@{}}C-10\\(\textcolor{red}{\small{10}}/\textcolor{blue}{\small{10K}})\end{tabular}} &
  \multicolumn{1}{l|}{\begin{tabular}[c]{@{}c@{}}C-100\\(\textcolor{red}{\small{100}}/\textcolor{blue}{\small{10K}})\end{tabular}} &
  \multicolumn{1}{l|}{\begin{tabular}[c]{@{}c@{}}Food\\(\textcolor{red}{\small{101}}/\textcolor{blue}{\small{25.3K}})\end{tabular}} &
  \multicolumn{1}{l|}{\begin{tabular}[c]{@{}c@{}}SUN\\(\textcolor{red}{\small{397}}/\textcolor{blue}{\small{19.9K}})\end{tabular}} &
  \multicolumn{1}{l|}{\begin{tabular}[c]{@{}c@{}}Cars\\(\textcolor{red}{\small{196}}/\textcolor{blue}{\small{8K}})\end{tabular}} &
  \multicolumn{1}{l|}{\begin{tabular}[c]{@{}c@{}}DTD\\(\textcolor{red}{\small{47}}/\textcolor{blue}{\small{1.9K}})\end{tabular}} &
  \multicolumn{1}{l|}{\begin{tabular}[c]{@{}c@{}}Caltech\\(\textcolor{red}{\small{101}}/\textcolor{blue}{\small{8.7K}})\end{tabular}} &
  \multicolumn{1}{l|}{\begin{tabular}[c]{@{}c@{}}Pets\\(\textcolor{red}{\small{37}}/\textcolor{blue}{\small{3.7K}})\end{tabular}} &
  \multicolumn{1}{l}{\begin{tabular}[c]{@{}c@{}}Places\\(\textcolor{red}{\small{365}}/\textcolor{blue}{\small{36.5K}})\end{tabular}} \\ \hline
CLIP (ViT-L/14) \cite{CLIP} &
  \multicolumn{1}{c|}{65.1} &
  \multicolumn{1}{c|}{87.6} &
  \multicolumn{1}{c|}{54.3} &
  \multicolumn{1}{c|}{86.8} &
  \multicolumn{1}{c|}{61.2} &
  \multicolumn{1}{c|}{65.1} &
  \multicolumn{1}{c|}{46.7} &
  \multicolumn{1}{c|}{83.4} &
  \multicolumn{1}{c|}{87.9} &
37.3
   \\
CuPL \cite{CuPL} &
  \multicolumn{1}{c|}{66.6} &
  \multicolumn{1}{c|}{86.6} &
  \multicolumn{1}{c|}{57.7} &
  \multicolumn{1}{c|}{89.0} &
  \multicolumn{1}{c|}{65.3} &
  \multicolumn{1}{c|}{63.9} &
  \multicolumn{1}{c|}{49.1} &
  \multicolumn{1}{c|}{90.5} &
  \multicolumn{1}{c|}{80.0} &
39.7
   \\\hdashline

Ours (descriptions) &
  \multicolumn{1}{c|}{71.3} &
  \multicolumn{1}{c|}{91.2} &
  \multicolumn{1}{c|}{65.3} &
  \multicolumn{1}{c|}{92.5} &
  \multicolumn{1}{c|}{68.8} &
  \multicolumn{1}{c|}{72.0} &
  \multicolumn{1}{c|}{55.1} &
  \multicolumn{1}{c|}{\colorbox{best}{\textbf{91.3}}} &
  \multicolumn{1}{c|}{85.0} &
  42.0 \\
Ours (class labels) &
  \multicolumn{1}{c|}{69.9} &
  \multicolumn{1}{c|}{91.8} &
  \multicolumn{1}{c|}{65.4} &
  \multicolumn{1}{c|}{92.0} &
  \multicolumn{1}{c|}{66.8} &
  \multicolumn{1}{c|}{74.3} &
  \multicolumn{1}{c|}{53.2} &
  \multicolumn{1}{c|}{88.5} &
  \multicolumn{1}{c|}{90.0} &
  41.3 \\
Ours (template) &
  \multicolumn{1}{c|}{69.0} &
  \multicolumn{1}{c|}{93.1} &
  \multicolumn{1}{c|}{65.8} &
  \multicolumn{1}{c|}{89.3} &
  \multicolumn{1}{c|}{64.5} &
  \multicolumn{1}{c|}{73.6} &
  \multicolumn{1}{c|}{52.3} &
  \multicolumn{1}{c|}{84.5} &
  \multicolumn{1}{c|}{87.9} &
  39.4 
  \\
Ours (combined) &
  \multicolumn{1}{c|}{\colorbox{best}{\textbf{73.4}}} &
  \multicolumn{1}{c|}{\colorbox{best}{\textbf{93.4}}} &
  \multicolumn{1}{c|}{\colorbox{best}{\textbf{70.2}}} &
  \multicolumn{1}{c|}{\colorbox{best}{\textbf{93.0}}} &
  \multicolumn{1}{c|}{\colorbox{best}{\textbf{70.6}}} &
  \multicolumn{1}{c|}{\colorbox{best}{\textbf{76.6}}} &
  \multicolumn{1}{c|}{\colorbox{best}{\textbf{58.0}}} &
  \multicolumn{1}{c|}{89.4} &
  \multicolumn{1}{c|}{\colorbox{best}{\textbf{90.9}}} &
   {\colorbox{best}{\textbf{43.4}}}
\end{tabular}
}
\end{table*}

The top-1 accuracy results are reported in Table \ref{tab:results} (see supplemental materials for top-5 accuracy results). As can be seen, our method consistently achieves the best accuracy when compared with baseline methods across all datasets. Using the combined class features yields the most promising results across the majority of datasets, with the exception of Caltech-101 \cite{caltech101}, where the best results were achieved using the class description features.

\begin{figure*}[t]
\centering
\includegraphics[width=.97\linewidth]{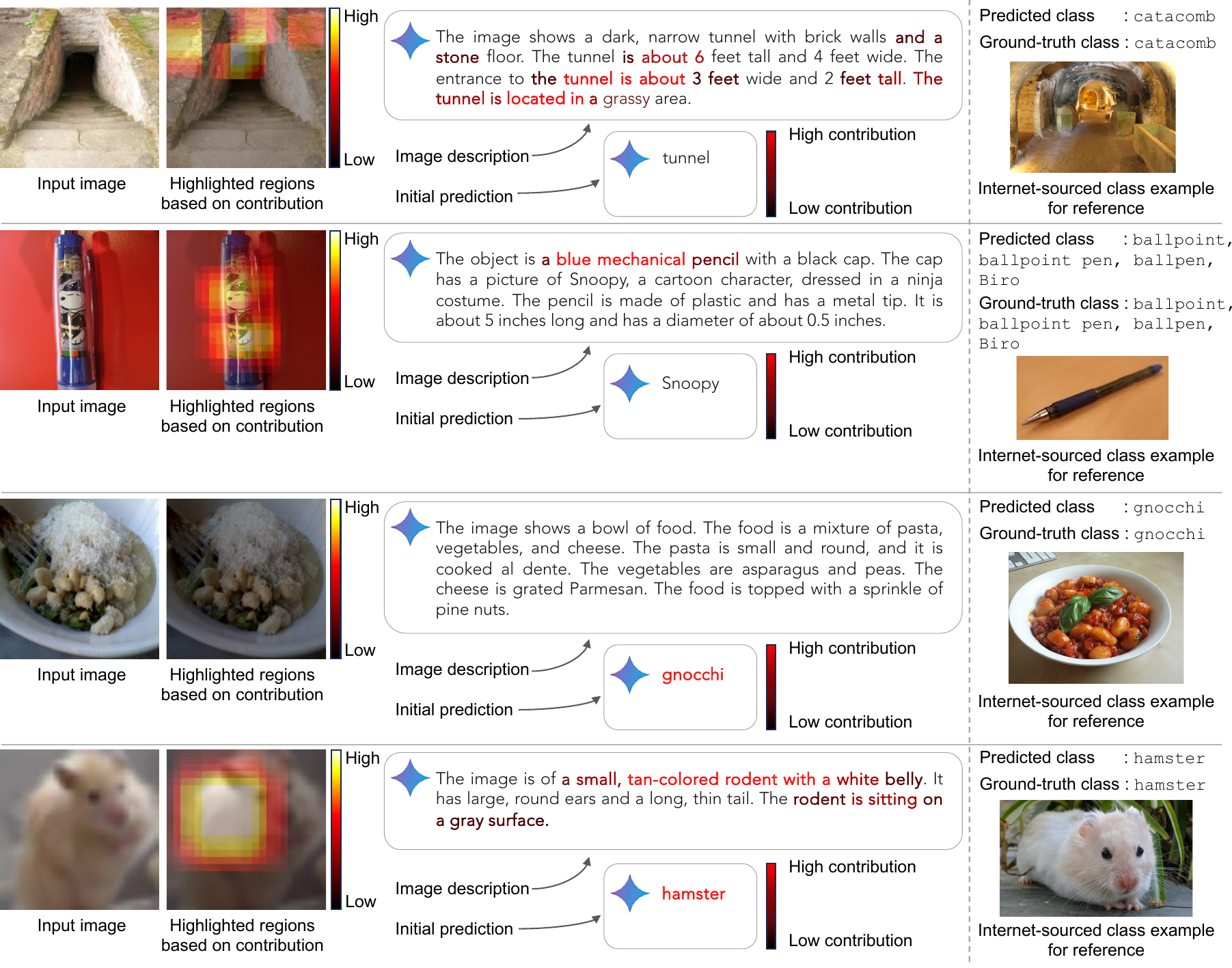}
\vspace{-2mm}
\caption{Input data highlighted based on its contribution to the final prediction. Examples are shown from the Places \cite{places} (first row), ImageNet \cite{imagenet} (second row), Food-101 \cite{food} (third row), and CIFAR-100 \cite{cifar} (last row) datasets.}
\label{fig:analysis}
\end{figure*}

Figure \ref{fig:analysis} shows visual examples, where we highlight parts of input modalities (image, initial prediction, and image description) based on their contribution to the final predicted class. To emphasize the significance of each input component in the final prediction, we employ a straightforward approach. Specifically, we utilize a 2D sliding kernel that traverses the image, masking out patches of the image. Subsequently, we measure the difference between the initial prediction and the prediction after masking to highlight areas of the image that contribute most significantly to the final prediction. Similarly, we apply this approach to the textual inputs. Given two distinct input texts -- namely, the initial prediction and the image description generated by the LLM (Gemini \cite{gemini}) -- we utilize a sliding kernel with a stride of one word. We mask out words that match the kernel and quantify their importance in our final prediction. 
As shown in Figure \ref{fig:analysis}, the three inputs collectively contribute to predicting the final class label. In some cases, one or two inputs exhibit a higher level of influence than the others, as demonstrated in the first, second, and third examples. 

It is worth mentioning that, while Gemini's initial predictions do not match the ground-truth class in the first and second examples, the predictions are contextually sensible. In the first example, Gemini's prediction was `{\texttt{tunnel}}' which, while not directly matching any class label in the Places dataset \cite{places}, conceptually aligns with the displayed `{\texttt{catacomb}}' image as an underground passage. Similarly, in the second example, Gemini's initial prediction was `{\texttt{Snoopy}}', which corresponds to the character drawn on the pen shown in the input image. However, `{\texttt{Snoopy}}' is not one of the ImageNet \cite{imagenet} class labels and the correct class of the shown image in second row of Figure \ref{fig:analysis} is `{\texttt{ballpoint pen}}'. This behavior of LLMs is the reason we cannot use them directly as image classifiers, because they sometimes do not restrict the output class to the provided list of target classes. However, such behavior might be beneficial to other classification tasks that are not restricted to a specific set of classes. Additional examples are provided in the supplemental materials.

\subsection{Ablation Studies}
\label{sec:ablations}

\begin{table*}[t]
\centering
\caption{\label{tab:ablation1}Ablation study on the impact of features used by our method on the classification accuracy. DF refers to the description feature, PF refers to the prediction feature, and IF refers to the image feature. In all datasets, we employed the best class feature as indicated in Table \ref{tab:results}. Specifically, we utilized the combined class feature for all datasets except for Caltech-101 \cite{caltech101}, where we opted for the class description feature. The best results are highlighted in \colorbox{best}{\textbf{bold}}.}
\vspace{-2mm}
\scalebox{0.85}{
\begin{tabular}{l|cccccccccc}
 &
  \multicolumn{9}{c}{\textbf{Dataset}} \\ \hline
\textbf{Method} &
  \multicolumn{1}{l|}{ImageNet} &
  \multicolumn{1}{l|}{C-10} &
  \multicolumn{1}{l|}{C-100} &
  \multicolumn{1}{l|}{Food} &
  \multicolumn{1}{l|}{SUN} &
  \multicolumn{1}{l|}{Cars} &
  \multicolumn{1}{l|}{DTD} &
  \multicolumn{1}{l|}{Caltech} &
  \multicolumn{1}{l|}{Pets} & 
  \multicolumn{1}{l}{Places} \\ \hline
Ours (DF) &
  \multicolumn{1}{c|}{58.6} &
  \multicolumn{1}{c|}{90.1} &
  \multicolumn{1}{c|}{64.5} &
  \multicolumn{1}{c|}{82.7} &
  \multicolumn{1}{c|}{49.0} &
  \multicolumn{1}{c|}{65.4} &
  \multicolumn{1}{c|}{49.8} &
  \multicolumn{1}{c|}{83.7} &
  \multicolumn{1}{c|}{48.3} &
30.1
   \\
Ours (PF) &
  \multicolumn{1}{c|}{55.7} &
  \multicolumn{1}{c|}{\colorbox{best}{\textbf{94.6}}} &
  \multicolumn{1}{c|}{73.2} &
  \multicolumn{1}{c|}{89.6} &
  \multicolumn{1}{c|}{60.9} &
  \multicolumn{1}{c|}{70.8} &
  \multicolumn{1}{c|}{57.7} &
  \multicolumn{1}{c|}{89.0} &
  \multicolumn{1}{c|}{87.1} &
36.1
   \\
Ours (DF and PF) &
    \multicolumn{1}{c|}{64.5} &
  \multicolumn{1}{c|}{94.4} &
  \multicolumn{1}{c|}{\colorbox{best}{\textbf{74.0}}} &
  \multicolumn{1}{c|}{89.8} &
  \multicolumn{1}{c|}{61.6} &
  \multicolumn{1}{c|}{71.6} &
  \multicolumn{1}{c|}{57.7} &
  \multicolumn{1}{c|}{89.3} &
  \multicolumn{1}{c|}{87.5} &
37.0
   \\
 Ours (DF and IF) &
    \multicolumn{1}{c|}{70.7} &
  \multicolumn{1}{c|}{90.4} &
  \multicolumn{1}{c|}{64.0} &
  \multicolumn{1}{c|}{90.8} &
  \multicolumn{1}{c|}{67.9} &
  \multicolumn{1}{c|}{71.4} &
  \multicolumn{1}{c|}{52.6} &
  \multicolumn{1}{c|}{90.9} &
  \multicolumn{1}{c|}{85.5} &
42.1
   \\
 Ours (PF and IF) &
    \multicolumn{1}{c|}{71.6} &
  \multicolumn{1}{c|}{92.0} &
  \multicolumn{1}{c|}{67.2} &
  \multicolumn{1}{c|}{92.2} &
  \multicolumn{1}{c|}{69.7} &
  \multicolumn{1}{c|}{74.1} &
  \multicolumn{1}{c|}{56.4} &
  \multicolumn{1}{c|}{91.1} &
  \multicolumn{1}{c|}{90.6} &
42.7
   \\
Ours (DF, PF, and IF) &
    \multicolumn{1}{c|}{\colorbox{best}{\textbf{73.4}}} &
  \multicolumn{1}{c|}{93.4} &
  \multicolumn{1}{c|}{70.2} &
  \multicolumn{1}{c|}{\colorbox{best}{\textbf{93.0}}} &
  \multicolumn{1}{c|}{\colorbox{best}{\textbf{70.6}}} &
  \multicolumn{1}{c|}{\colorbox{best}{\textbf{76.6}}} &
  \multicolumn{1}{c|}{\colorbox{best}{\textbf{58.0}}} &
  \multicolumn{1}{c|}{\colorbox{best}{\textbf{91.3}}} &
  \multicolumn{1}{c|}{\colorbox{best}{\textbf{90.9}}} &
\colorbox{best}{\textbf{43.4}}
\end{tabular}
}
\end{table*}

We conducted a series of ablation studies to explore different versions of our method and investigate the impact of each individual feature, different fusion approaches between the features, and different cross-modal embedding models. Table \ref{tab:ablation1} presents the results of our method using solely the encoded image description, referred to as the description feature (DF). We also report the results obtained by using the encoding of the initial class predictions, termed as the prediction feature (PF), as well as using both DF and PF concurrently as inputs. Additionally, Table \ref{tab:ablation1} shows the results of using image feature (IF) alongside DF or PF as inputs, and finally, we present the results when leveraging all available input features (i.e., DF, PF, and IF). 

From the results in Table \ref{tab:ablation1}, it is clear that incorporating all three features (i.e., DF, PF, and IF) yields the best performance across most datasets, except for the CIFAR datasets \cite{cifar}. This discrepancy may arise from the low resolution of CIFAR images (originally 32$\times$32 pixels), where utilizing the image feature (IF) may degrade accuracy compared to using only DF and PF.

\begin{table}[t]
    \centering
    \caption{\label{tab:ablation2}Ablation study on various fusion approaches using 5,000 images randomly selected from the ImageNet dataset \cite{imagenet}. The best results are highlighted in \colorbox{best}{\textbf{bold}}.}
        \centering
        \scalebox{0.8}{
            \begin{tabular}{l|ccc}
                & \multicolumn{3}{c}{\textbf{CLIP model} \cite{CLIP}} \\ \hline
                \textbf{Fusion Approach} & ViT-L/14 & ViT-B/32  & ViT-B/16 \\ \hline
                Max similarity           &  57.6                              &  57.8    & 57.9      \\
                Avg similarity           &  66.1                              &  65.5    & 65.8      \\
                Avg feature              &  \colorbox{best}{\textbf{72.5}}    &  65.7    & 68.5         
            \end{tabular}
        }
\end{table}

\begin{figure*}[t]
\centering
\includegraphics[width=.96\linewidth]{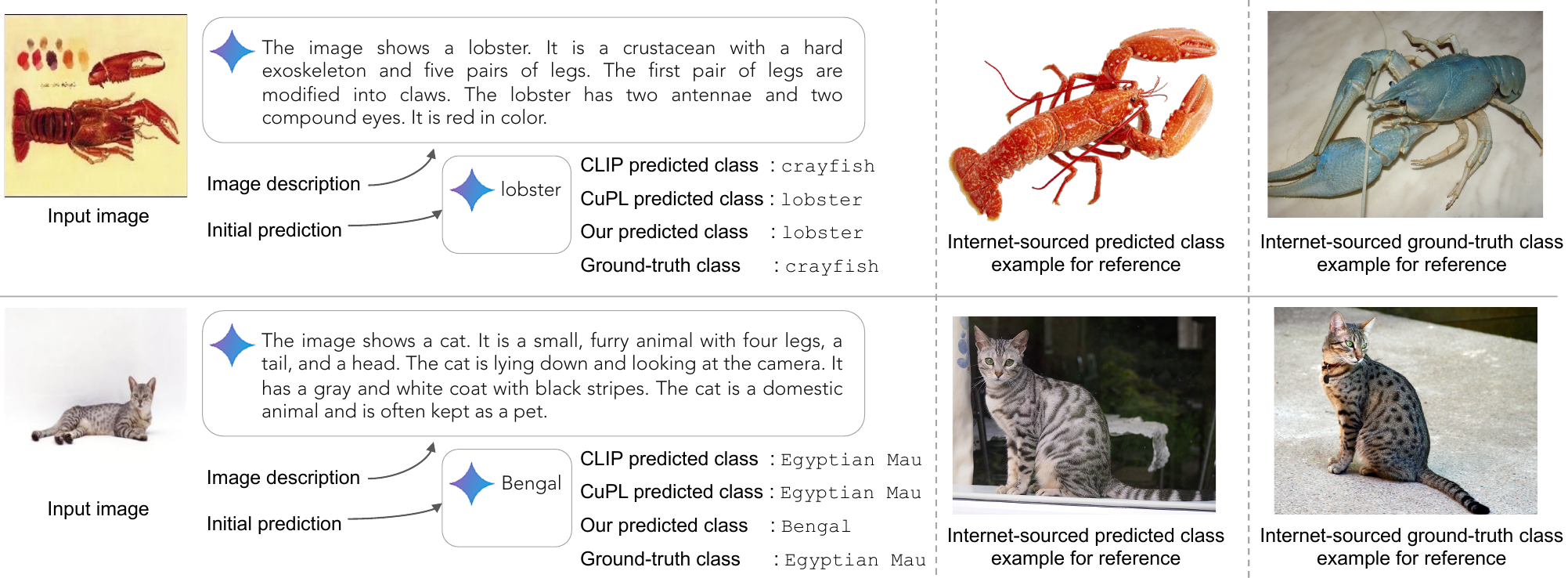}

\caption{Failure examples of our method, where the initial prediction (and the image description in the first example) adversely influenced our final decision. Results are shown for the Caltech-101 dataset \cite{caltech101} (first row) and the Pets dataset \cite{pets} (second row).}
\label{fig:failure_cases}
\end{figure*}

\begin{table}[t]
\centering
\caption{\label{tab:ablation3}Top-1 accuracy comparing our method with alternative approaches for aligning Gemini's class predictions with dataset labels. These results are reported on 5,000 randomly selected images from the ImageNet dataset \cite{imagenet}. Best result is highlighted in \colorbox{best}{\textbf{yellow}}}
\scalebox{0.8}{
\begin{tabular}{cccc}
\toprule
L/14 & B/16 & B/32 & DistilBERT \\
\hline
52.1  &   52.0  &   55.6  &    43.7 \\
\toprule
RoBERTa & ROUGE-N-F1 & ROUGE-F1 & Ours \\ 
\hline
28.8  &  54.8 & 54.9 &  \colorbox{best}{\textbf{70.2}} \\
\bottomrule
\end{tabular}}
\end{table}

Table \ref{tab:ablation2} shows the results of our second set of ablation studies, where we report the results for 5,000 randomly selected images from the ImageNet dataset \cite{imagenet}. We explore the use of different cross-modal embedding models (CLIP [ViT-L/14], CLIP [ViT-B/32], and CLIP [ViT-B/16]) \cite{CLIP}, and additionally investigate different fusion approaches. Rather than using the mean feature vector of our input features (DF, PF, and IF), we calculated the similarity between each input feature separately and the dataset class label features. Subsequently, we fused the similarity scores to generate a single similarity score for each class label in each dataset. We explored two fusion methods: averaging and taking the maximum for each class label. As shown, averaging our three features (IF, DF, PF) yields the best results. Generally, averaging is more robust and less sensitive to outliers than taking the maximum, and hence, more representative. 

As shown in Tables~\ref{tab:results} and \ref{tab:ablation1}, encouraging results were demonstrated by utilizing the encoded features from the LLM (i.e., Gemini \cite{gemini}) for zero-shot image classification. Based on these results, one might argue for the direct utilization of Gemini's class prediction, aiming to match a specific class label from the dataset. However, in several cases, Gemini's response does not precisely match one of the class labels. For example, if a ground-truth class label is `{\ttfamily{cat}}', Gemini's response might be `{\ttfamily{The image class is cat}}' (see Fig.~\ref{fig:intro}). This discrepancy motivated us to report results of using only Gemini prediction. 

Specifically, we present additional results from early experiments aimed at utilizing Gemini's predictions to precisely match one of the class labels in the given dataset. We randomly selected 5,000 images from ImageNet \cite{imagenet} for evaluation. While our method, offers a practical way of utilizing Gemini's predictions, we also present the results of some alternative approaches aimed at precisely identifying one of the dataset class labels, rather than solely relying on the class prediction text generated by Gemini.

Table \ref{tab:ablation3} show the results on the 5,000 images from ImageNet \cite{imagenet} of our main method and alternatives that utilize Gemini's class prediction to conduct similarity matching with the target dataset class labels. Specifically, we report the results of encoding Gemini's class prediction using an open-vocabulary language model and measuring the similarity with the encoded class label features. Here, we show the results of using CLIP (ViT-L/14, ViT-B/32, ViT-B/16) \cite{CLIP}, DistilBERT \cite{DistilBERT}, and RoBERTa \cite{RoBERTa}. 

In addition, we explored classical text similarity metrics -- namely, ROUGE-N-F1 and ROUGE-F1 \cite{ROUGE} -- rather than encoding both Gemini's prediction and class labels using an open-vocabulary encoding model. As shown in Table \ref{tab:ablation3}, our method, which utilizes Gemini's class prediction as one of the input features, achieves the best results when compared with the alternative approaches.

\subsection{Computation resources}
\label{sec:resources}
Our method relies on a multimodal LLM and cross-modal encoders. The cross-modal encoding takes around 15 ms to encode an image or text on an NVIDIA V100 GPU, while the LLM can be accessed through: 1) Cloud API calls, which do not require local resources to load the model, or 2) loading the model locally for processing, which requires an estimated 16 GPUs/TPUs with approximately 256 GB of memory. Each LLM query takes roughly 700 ms to process. The LLM model is the most intensive operation (as discussed in Section~\ref{sec:limitations}), but it can be accelerated using multi-threading.

\section{Limitations}
\label{sec:limitations}

Our method introduces a new approach by leveraging multimodal LLMs to enhance the accuracy of zero-shot image classification. However, it is important to acknowledge that there are still some limitations inherent in our proposed method. Since our method relies on a multimodal LLM to generate the required features (i.e., DF and PF), there may be potential constraints when running on devices with limited computational power, and it may consume more time compared to other methods. Nevertheless, we believe that advancements in LLMs will lead to models that can run efficiently on lower computational power. This would enable broader accessibility and applicability of such models, such as Gemini \cite{gemini}, GPT \cite{GPT}, and LLaMA \cite{LLaMA}.

Our method fails in some cases. Figure \ref{fig:failure_cases} shows examples of failure cases, where our method misclassify the input image. While the initial prediction and image description features generally enhance classification accuracy, as demonstrated in Table \ref{tab:ablation1}, they can sometimes lead to misclassifications. In the first example in Figure \ref{fig:failure_cases}, both the image description and initial prediction suggest that the image show a `{\ttfamily{lobster}}', whereas it actually shows a `{\ttfamily{crayfish}}'. Similarly, in the second example, the image description lacks specific features of the cat, while the initial prediction suggests the `{\ttfamily{Bengal}}' class label, whereas the actual class label is `{\ttfamily{Egyptian Mau}}'.

\section{Conclusion}
\label{sec:conclusion}
We proposed a simple yet effective zero-shot image classification method leveraging multimodal LLMs. Our approach uses an LLM to generate a description of an input image and make an initial class prediction. We then fuse encoded features from the image description, the initial LLM prediction, and the image itself. This straightforward method significantly improves zero-shot classification accuracy compared to existing techniques under the same experimental conditions. Future work will focus on deploying this approach with on-device multimodal LLMs to reduce the computational and communication overhead of cloud-based models.

\clearpage
\maketitlesupplementary
\appendix

\section{Additional Details}
\label{supp-details}

In the main paper, we presented our method for zero-shot image classification. Here we give an example of implementing our method, including both classifier construction and the inference process, as shown in Code~\ref{code}.

\begin{figure*}[!t]
\centering
\includegraphics[width=\linewidth]{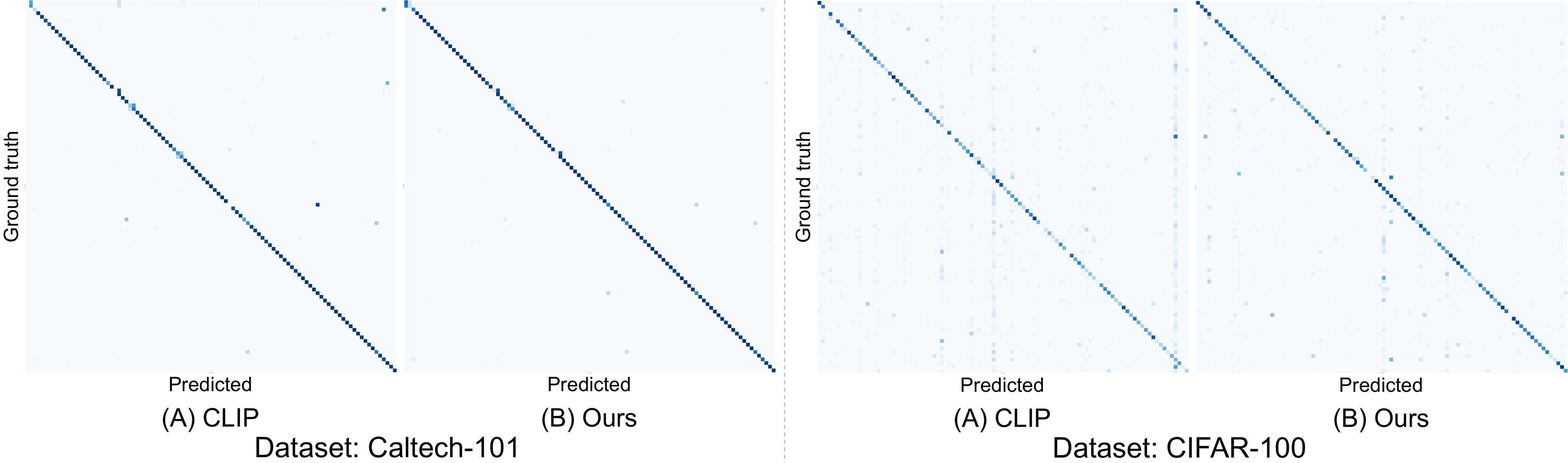}
\vspace{-3pt}
\caption{Confusion matrices for zero-shot image classification results of (A) CLIP (ViT-L/14) \cite{CLIP} and (B) our method on the Caltech-101 \cite{caltech101} and CIFAR-100 \cite{cifar} datasets.}
\label{fig:confusion_matrirx}
\end{figure*}

\begin{table*}[t]
\centering
\caption{\label{tab:results2}Top-5 classification accuracy of CLIP \cite{CLIP}, CuPL \cite{CuPL}, and our method on the following datasets: ImageNet \cite{imagenet}, CIFAR-10 (C-10) \cite{cifar}, CIFAR-100 (C-100) \cite{cifar}, Food-101 \cite{food}, SUN397 \cite{sun1, sun2}, Cars \cite{cars}, DTD \cite{dtd}, Caltech-101 \cite{caltech101}, Pets \cite{pets}, and Places \cite{places}. We report our results with the following class label features: 1) class descriptions, 2) class labels, 3) the template ``{\texttt{A photo of \{class\}}}'', and 4) combined features of (1-3). The best results are highlighted in \colorbox{best}{\textbf{yellow}}.}
\scalebox{0.8}{
\begin{tabular}{l|cccccccccc}
 &
  \multicolumn{9}{c}{\textbf{Dataset}} \\ \hline
\textbf{Method} &
  \multicolumn{1}{l|}{ImageNet} &
  \multicolumn{1}{l|}{C-10} &
  \multicolumn{1}{l|}{C-100} &
  \multicolumn{1}{l|}{Food} &
  \multicolumn{1}{l|}{SUN} &
  \multicolumn{1}{l|}{Cars} &
  \multicolumn{1}{l|}{DTD} &
  \multicolumn{1}{l|}{Caltech} &
  \multicolumn{1}{l|}{Pets} & 
  \multicolumn{1}{l}{Places} \\ \hline
CLIP (ViT-L/14) \cite{CLIP} &
  \multicolumn{1}{c|}{88.4} &
  \multicolumn{1}{c|}{98.5} &
  \multicolumn{1}{c|}{77.0} &
  \multicolumn{1}{c|}{97.8} &
  \multicolumn{1}{c|}{89.1} &
  \multicolumn{1}{c|}{93.7} &
  \multicolumn{1}{c|}{72.8} &
  \multicolumn{1}{c|}{95.2} &
  \multicolumn{1}{c|}{96.6} &
64.5
  \\
CuPL \cite{CuPL} &
  \multicolumn{1}{c|}{91.0} &
  \multicolumn{1}{c|}{98.1} &
  \multicolumn{1}{c|}{79.3} &
  \multicolumn{1}{c|}{98.3} &
  \multicolumn{1}{c|}{92.1} &
  \multicolumn{1}{c|}{94.2} &
  \multicolumn{1}{c|}{77.7} &
  \multicolumn{1}{c|}{99.8} &
  \multicolumn{1}{c|}{96.2} &
68.8
  \\
\hdashline

Ours (class descriptions) &
    \multicolumn{1}{c|}{92.7} &
  \multicolumn{1}{c|}{99.3} &
  \multicolumn{1}{c|}{85.4} &
  \multicolumn{1}{c|}{98.9} &
  \multicolumn{1}{c|}{93.8} &
  \multicolumn{1}{c|}{97.6} &
  \multicolumn{1}{c|}{81.0} &
  \multicolumn{1}{c|}{\colorbox{best}{\textbf{99.9}}} &
  \multicolumn{1}{c|}{96.9} &
  70.7
  \\
Ours (class labels) &
    \multicolumn{1}{c|}{89.9} &
  \multicolumn{1}{c|}{99.2} &
  \multicolumn{1}{c|}{84.3} &
  \multicolumn{1}{c|}{98.8} &
  \multicolumn{1}{c|}{91.1} &
  \multicolumn{1}{c|}{97.5} &
  \multicolumn{1}{c|}{77.6} &
  \multicolumn{1}{c|}{98.9} &
  \multicolumn{1}{c|}{98.6} &
  67.5
  \\
Ours (template) &
  \multicolumn{1}{c|}{89.8} &
  \multicolumn{1}{c|}{\colorbox{best}{\textbf{99.6}}} &
  \multicolumn{1}{c|}{84.8} &
  \multicolumn{1}{c|}{97.9} &
  \multicolumn{1}{c|}{90.1} &
  \multicolumn{1}{c|}{97.5} &
  \multicolumn{1}{c|}{77.2} &
  \multicolumn{1}{c|}{96.6} &
  \multicolumn{1}{c|}{97.3} &
  65.3
  \\
Ours (combined) & 
  \multicolumn{1}{c|}{\colorbox{best}{\textbf{93.0}}} &
  \multicolumn{1}{c|}{\colorbox{best}{\textbf{99.6}}} &
  \multicolumn{1}{c|}{\colorbox{best}{\textbf{88.5}}} &
  \multicolumn{1}{c|}{\colorbox{best}{\textbf{99.0}}} &
  \multicolumn{1}{c|}{\colorbox{best}{\textbf{94.4}}} &
  \multicolumn{1}{c|}{\colorbox{best}{\textbf{97.9}}} &
  \multicolumn{1}{c|}{\colorbox{best}{\textbf{83.8}}} &
  \multicolumn{1}{c|}{\colorbox{best}{\textbf{99.9}}} &
  \multicolumn{1}{c|}{\colorbox{best}{\textbf{99.5}}} &
  \colorbox{best}{\textbf{70.9}}
  \end{tabular}
}
\end{table*}

\begin{table*}[t]
\centering
\caption{\label{tab:kappa-score}Cohen's Kappa score of CLIP \cite{CLIP}, CuPL \cite{CuPL}, and our method on the following datasets: ImageNet \cite{imagenet}, CIFAR-10 (C-10) \cite{cifar}, CIFAR-100 (C-100) \cite{cifar}, Food-101 \cite{food}, SUN397 \cite{sun1, sun2}, Cars \cite{cars}, DTD \cite{dtd}, Caltech-101 \cite{caltech101}, Pets \cite{pets}, and Places \cite{places}. We report our results with the following class label features: 1) class descriptions, 2) class labels, 3) the template ``{\texttt{A photo of \{class\}}}'', and 4) combined features of (1-3). The best results are highlighted in \colorbox{best}{\textbf{yellow}}.}
\scalebox{0.75}{
\begin{tabular}{l|cccccccccc}
 &
  \multicolumn{9}{c}{\textbf{Dataset}} \\ \hline
\textbf{Method} &
  \multicolumn{1}{l|}{ImageNet} &
  \multicolumn{1}{l|}{C-10} &
  \multicolumn{1}{l|}{C-100} &
  \multicolumn{1}{l|}{Food} &
  \multicolumn{1}{l|}{SUN} &
  \multicolumn{1}{l|}{Cars} &
  \multicolumn{1}{l|}{DTD} &
  \multicolumn{1}{l|}{Caltech} &
  \multicolumn{1}{l|}{Pets} & 
  \multicolumn{1}{l}{Places} \\ \hline

CLIP (ViT-L/14) \cite{CLIP} &
  \multicolumn{1}{c|}{0.651} &
  \multicolumn{1}{c|}{0.862} &
  \multicolumn{1}{c|}{0.539} &
  \multicolumn{1}{c|}{0.867} &
  \multicolumn{1}{c|}{0.611} &
  \multicolumn{1}{c|}{0.656} &
  \multicolumn{1}{c|}{0.452} &
  \multicolumn{1}{c|}{0.830} &
  \multicolumn{1}{c|}{0.835} &
  0.372
  \\
CuPL \cite{CuPL} &
  \multicolumn{1}{c|}{0.665} &
  \multicolumn{1}{c|}{0.851} &
  \multicolumn{1}{c|}{0.573} &
  \multicolumn{1}{c|}{0.889} &
  \multicolumn{1}{c|}{0.652} &
  \multicolumn{1}{c|}{0.637} &
  \multicolumn{1}{c|}{0.480} &
  \multicolumn{1}{c|}{0.902} &
  \multicolumn{1}{c|}{0.795} &
  0.395
  \\
\hdashline

Ours (class descriptions) &
  \multicolumn{1}{c|}{0.713} &
  \multicolumn{1}{c|}{0.902} &
  \multicolumn{1}{c|}{0.650} &
  \multicolumn{1}{c|}{0.924} &
  \multicolumn{1}{c|}{0.687} &
  \multicolumn{1}{c|}{0.718} &
  \multicolumn{1}{c|}{0.541} &
  \multicolumn{1}{c|}{\colorbox{best}{\textbf{0.911}}} &
  \multicolumn{1}{c|}{0.846} &
  0.418
  \\
Ours (class labels) &
  \multicolumn{1}{c|}{0.699} &
  \multicolumn{1}{c|}{0.909} &
  \multicolumn{1}{c|}{0.650} &
  \multicolumn{1}{c|}{0.920} &
  \multicolumn{1}{c|}{0.667} &
  \multicolumn{1}{c|}{0.742} &
  \multicolumn{1}{c|}{0.522} &
  \multicolumn{1}{c|}{0.882} &
  \multicolumn{1}{c|}{0.897} &
  0.411
  \\
Ours (template) &
  \multicolumn{1}{c|}{0.709} &
  \multicolumn{1}{c|}{0.925} &
  \multicolumn{1}{c|}{0.675} &
  \multicolumn{1}{c|}{0.919} &
  \multicolumn{1}{c|}{0.668} &
  \multicolumn{1}{c|}{0.747} &
  \multicolumn{1}{c|}{0.545} &
  \multicolumn{1}{c|}{0.874} &
  \multicolumn{1}{c|}{0.892} &
  0.412
  \\
Ours (combined) & 
  \multicolumn{1}{c|}{\colorbox{best}{\textbf{0.734}}} &
  \multicolumn{1}{c|}{\colorbox{best}{\textbf{0.927}}} &
  \multicolumn{1}{c|}{\colorbox{best}{\textbf{0.699}}} &
  \multicolumn{1}{c|}{\colorbox{best}{\textbf{0.929}}} &
  \multicolumn{1}{c|}{\colorbox{best}{\textbf{0.706}}} &
  \multicolumn{1}{c|}{\colorbox{best}{\textbf{0.764}}} &
  \multicolumn{1}{c|}{\colorbox{best}{\textbf{0.571}}} &
  \multicolumn{1}{c|}{0.890} &
  \multicolumn{1}{c|}{\colorbox{best}{\textbf{0.907}}} &
  \colorbox{best}{\textbf{0.432}}
  \end{tabular}
}
\end{table*}

In the main paper, we visualize examples that highlight the important parts of the inputs contributing to the final predicted class label. We described the approach of sequentially masking out patches from the image and comparing the predicted class with the prediction obtained using the entire unmasked image. Similarly, we follow the same approach for text input by sliding a kernel, masking out words, and comparing the predicted class with our original prediction using inputs without any masking. We used a 2D kernel of size 50$\times$50 pixels with a stride of 10 pixels. If there are no highlighted regions in the image due to the small size of the kernel, we enlarge it by 50 until we reach a kernel size of 200$\times$200 pixels.

For the text kernel, we start with a kernel width of 3 words. If none of the words are highlighted, we reduce it by 1 until we use a 1-word kernel sliding over the text. Each prediction was made using the three inputs: the image, initial prediction, and image description, with one of them having masked out patches or words.

\begin{figure*}[t]
\begin{lstlisting}[caption={Example Python implementation of our method. In this example, we utilize the combined class feature.}, label={code}, style=tfstyle]
import tensorflow as tf
import cross_modal_encoder as encoder  # for example CLIP
import llm  # for example Gemini Pro
from fixed_prompts import classification_p, description_p, class_ps  # see Table 4.

def create_classifer(class_names, k=50):
  '''Constructs zero-shot image classifier.
  Args:
    class_names: A list of class names.
    k: Number of class descriptions to be generated by the LLM.
  Returns:
    A zero-shot image classification model.
  '''
  assert k >= len(class_ps)
  assert k % len(class_ps) == 0
  weights = []
  for class_name in class_names:
    class_name_feature = encoder.encode_text(class_name)
    template_feature = encoder.encode_text(f"A photo of {class_name}")
    llm_class_description = tf.zeros((1, encoder.output_feature_length))
    for _ in range(k // len(class_ps)):
      for class_p in class_ps:
        llm_class_feature = llm.process(class_p.format(class_name), temperature=0.99)
        llm_class_description += encoder.encode_text(llm_class_feature)
    llm_class_description /= k
    class_feature = class_name_feature + template_feature + llm_class_description
    normalized_class_feature = class_feature / tf.norm(class_feature)
    weights.append(tf.squeeze(normalized_class_feature))
  model = {"weights": tf.transpose(tf.convert_to_tensor(weights)), 
           "class_names": class_names}
  return model


def classify(image, classifier):
  '''Performs zero-shot image classification.
  Args:
    image: Input testing image.
    classifier: A zero-shot classification model generated by create_classifier function.
  Returns:
    Predicted class name.
  '''
  image_feature = encoder.encode_image(image)
  image_feature /= tf.norm(image_feature)
  initial_prediction = llm.process([classification_p, image], temperature=0)
  prediction_feature = encoder.encode_text(initial_prediction)
  prediction_feature /= tf.norm(prediction_feature)
  image_description = llm.process([description_p, image], temperature=0)
  description_feature = encoder.encode_text(image_description)
  description_feature /= tf.norm(description_feature)
  query_feature = image_feature + prediction_feature + description_feature
  query_feature /= tf.norm(query_feature)
  index = tf.argmax(tf.linalg.matmul(query_feature, classifier["weights"]))
  return classifier["class_names"][index.numpy().squeeze()]
\end{lstlisting}
  \vspace{-8pt}
\end{figure*}

\section{Additional Results Using Gemini}
\label{supp-results}
In this section, we provide supplementary results to those presented in the main paper. Figure \ref{fig:confusion_matrirx} shows the confusion matrix for CLIP (ViT-L/14) \cite{CLIP} and our method across two datasets (Caltech-101 \cite{caltech101} and CIFAR-100 \cite{cifar}). The shown results demonstrate that our method enhances classification accuracy and reduces misclassification rates.

\begin{figure*}[!t]
\centering
\includegraphics[width=\linewidth]{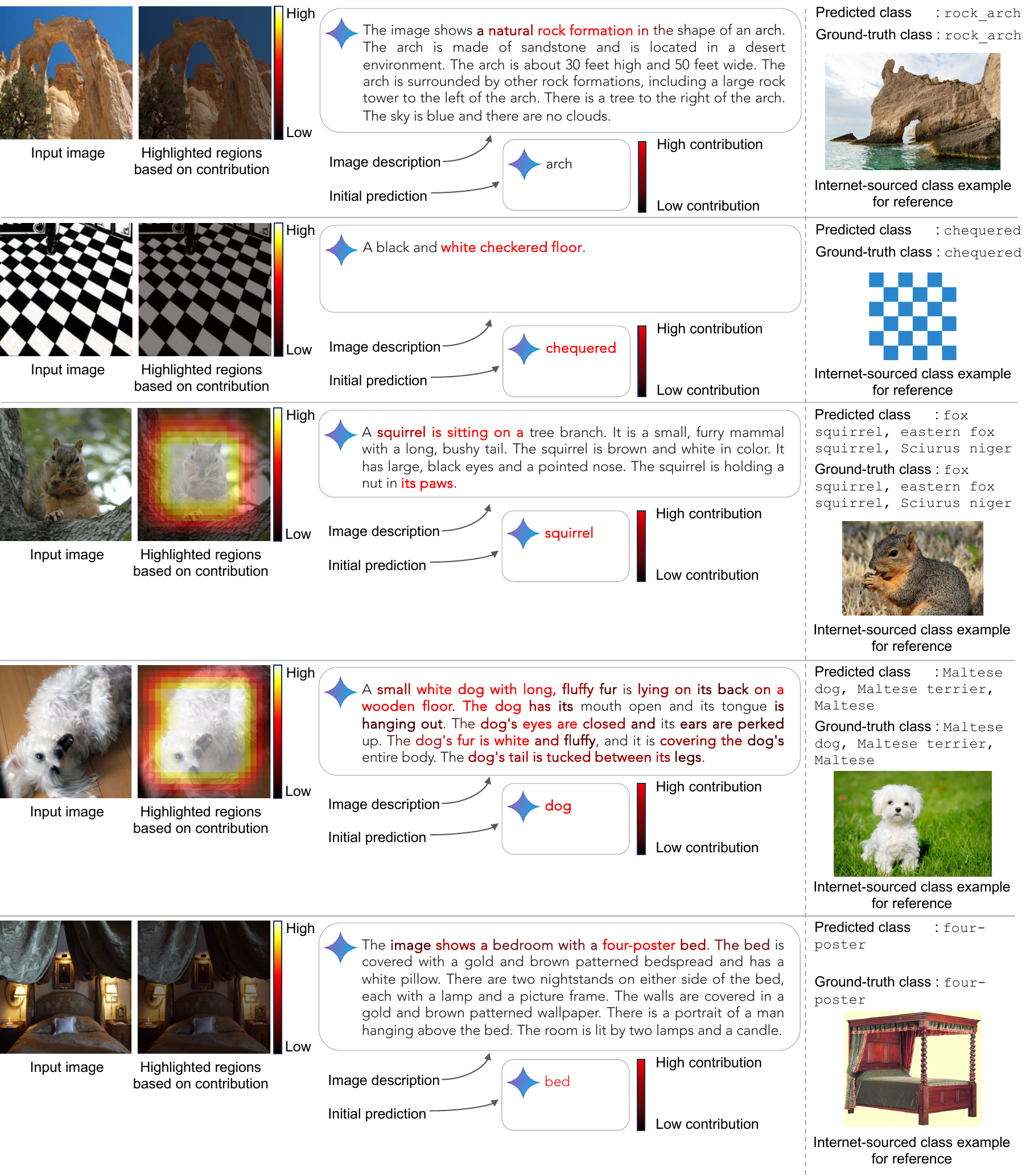}
\vspace{-3pt}
\caption{Additional examples demonstrating the influence of input data on final predictions. Examples are provided from the following datasets: SUN397 \cite{sun1, sun2} (first row), DTD \cite{dtd} (second row), and ImageNet \cite{imagenet} (last three rows).}
\label{fig:analysis_supp}
\end{figure*}

In the main paper, we reported the top-1 classification accuracy on several datasets \cite{imagenet, caltech101, food, cars, cifar, dtd, places, sun1, sun2, pets}. Table \ref{tab:results2} presents the top-5 classification accuracy of our method compared to prior work, while Table \ref{tab:kappa-score} shows the Cohen's kappa coefficient. As can be seen, our method achieves a notable improvement while remaining simple and easy to implement.

Lastly, Figure \ref{fig:analysis_supp} shows additional visual examples, where we highlight the most significant contributors from input parts that influence the final predictions of our method.

\section{Results Using GPT}
\label{supp-gpt-results}

Our primary goal in this paper is not to compare different LLMs, but to evaluate zero-shot classification methods under a fair setting using the same vision and language backbones. Nonetheless, to assess the generalization of our method across different LLMs, we conducted experiments using GPT-4 \cite{GPT} as the LLM backbone. Results on the ImageNet validation set \cite{imagenet} are reported in Table~\ref{tab:rebuttal}, alongside CuPL \cite{CuPL}, using identical backbones (ViT-L/14 for CLIP and GPT for the LLM). We also compare with recent methods \cite{menon2022visual, mirza2024meta, parashar2024neglected} that adopt similar experimental setups. As shown, our method continues to achieve state-of-the-art results even with GPT-4. Notably, using only the raw GPT output---used to construct the `initial prediction feature'---achieves 61.1\% on ImageNet (ROUGE-F1), which underscores the necessity of our approach: LLM-generated outputs do not always align precisely with class labels.

\begin{table}[t]
\centering
\caption{\label{tab:rebuttal}Results w/ GPT LLM \cite{GPT} and CLIP (ViT-L/14) \cite{CLIP} on ImageNet \cite{imagenet}. The best results are highlighted in \colorbox{best}{\textbf{yellow}}.}
\vspace{-3mm}
\scalebox{0.85}{
\begin{tabular}{l|c}
Method                      & Accuracy (\%)\\ \hline
CuPL \cite{CuPL}               & 75.7     \\ \hline
CuPL \cite{CuPL} + REAL-Prompt \cite{parashar2024neglected} & 75.9     \\ \hline
DCLIP \cite{menon2022visual}                    & 74.5     \\ \hline
DCLIP + REAL-Prompt \cite{parashar2024neglected}        & 74.9     \\ \hline
MPVR (GPT+TEMP)    \cite{mirza2024meta}         & 76.8     \\ \hdashline
Ours                        &    \colorbox{best}{\textbf{78.4}}
\end{tabular}
}\vspace{-6mm}
\end{table}

\section{Broader Impact}
\label{supp-broader-impact}
Our work introduces a method for zero-shot image classification that leverages the power of multimodal LLMs not only during the classifier model construction phase but also at inference time. We achieve this by generating comprehensive textual representations directly from input images. These representations are then combined with the input images for classification, resulting in a significant enhancement in accuracy. 

Importantly, our approach eliminates the need for dataset-specific prompt engineering, as commonly required in prior approaches, thereby simplifying the implementation process and enhancing accessibility -- effectively acting as a plug-and-play solution. By removing the requirement for dataset-specific customization, our method offers a straightforward and user-friendly approach to zero-shot image classification, making it more accessible to a broader range of users.

By demonstrating its effectiveness across diverse datasets, we illustrate the utility of our method for robust and generalizable real-world computer vision systems reliant on image classification, eliminating the need for dataset-specific training, tuning, or prompt engineering. This approach holds promise for simplifying the deployment of image classification systems and advancing the field of computer vision.

{
    \small
    \bibliographystyle{ieeenat_fullname}
    \bibliography{main}

\begin{thebibliography}{29}
\providecommand{\natexlab}[1]{#1}
\providecommand{\url}[1]{\texttt{#1}}
\expandafter\ifx\csname urlstyle\endcsname\relax
  \providecommand{\doi}[1]{doi: #1}\else
  \providecommand{\doi}{doi: \begingroup \urlstyle{rm}\Url}\fi

\bibitem[Allingham et~al.(2023)Allingham, Ren, Dusenberry, Gu, Cui, Tran, Liu, and Lakshminarayanan]{allingham2023simple}
James~Urquhart Allingham, Jie Ren, Michael~W Dusenberry, Xiuye Gu, Yin Cui, Dustin Tran, Jeremiah~Zhe Liu, and Balaji Lakshminarayanan.
\newblock A simple zero-shot prompt weighting technique to improve prompt ensembling in text-image models.
\newblock In \emph{ICML}, 2023.

\bibitem[Bossard et~al.(2014)Bossard, Guillaumin, and Van~Gool]{food}
Lukas Bossard, Matthieu Guillaumin, and Luc Van~Gool.
\newblock Food-101 -- mining discriminative components with random forests.
\newblock In \emph{ECCV}, 2014.

\bibitem[Brown et~al.(2020)Brown, Mann, Ryder, Subbiah, Kaplan, Dhariwal, Neelakantan, Shyam, Sastry, Askell, et~al.]{GPT}
Tom Brown, Benjamin Mann, Nick Ryder, Melanie Subbiah, Jared~D Kaplan, Prafulla Dhariwal, Arvind Neelakantan, Pranav Shyam, Girish Sastry, Amanda Askell, et~al.
\newblock Language models are few-shot learners.
\newblock In \emph{NeurIPS}, 2020.

\bibitem[Cimpoi et~al.(2014)Cimpoi, Maji, Kokkinos, Mohamed, and Vedaldi]{dtd}
Mircea Cimpoi, Subhransu Maji, Iasonas Kokkinos, Sammy Mohamed, and Andrea Vedaldi.
\newblock Describing textures in the wild.
\newblock In \emph{CVPR}, 2014.

\bibitem[Deng et~al.(2009)Deng, Dong, Socher, Li, Li, and Fei-Fei]{imagenet}
Jia Deng, Wei Dong, Richard Socher, Li-Jia Li, Kai Li, and Li Fei-Fei.
\newblock {ImageNet}: A large-scale hierarchical image database.
\newblock In \emph{CVPR}, 2009.

\bibitem[Desai and Johnson(2021)]{VirTex}
Karan Desai and Justin Johnson.
\newblock {VirTex}: Learning visual representations from textual annotations.
\newblock In \emph{CVPR}, 2021.

\bibitem[et~al.(2024{\natexlab{a}})]{mirza2024meta}
Mirza et al.
\newblock Meta-prompting for automating zero-shot visual recognition with {LLM}s.
\newblock In \emph{ECCV}, 2024{\natexlab{a}}.

\bibitem[et~al.(2024{\natexlab{b}})]{parashar2024neglected}
Parashar et al.
\newblock The neglected tails in vision-language models.
\newblock In \emph{CVPR}, 2024{\natexlab{b}}.

\bibitem[Fei-Fei et~al.(2004)Fei-Fei, Fergus, and Perona]{caltech101}
Li Fei-Fei, Rob Fergus, and Pietro Perona.
\newblock Learning generative visual models from few training examples: An incremental bayesian approach tested on 101 object categories.
\newblock In \emph{CVPR workshops}, 2004.

\bibitem[Frome et~al.(2013)Frome, Corrado, Shlens, Bengio, Dean, Ranzato, and Mikolov]{DeViSE}
Andrea Frome, Greg~S Corrado, Jon Shlens, Samy Bengio, Jeff Dean, Marc'Aurelio Ranzato, and Tomas Mikolov.
\newblock {DeViSE}: A deep visual-semantic embedding model.
\newblock In \emph{NeurIPS}, 2013.

\bibitem[{Gemini Team Google}(2023)]{gemini}
{Gemini Team Google}.
\newblock Gemini: A family of highly capable multimodal models.
\newblock \emph{arXiv preprint arXiv:2312.11805}, 2023.

\bibitem[Guo et~al.(2023)Guo, Zhang, Qiu, Ma, Miao, He, and Cui]{CALIP}
Ziyu Guo, Renrui Zhang, Longtian Qiu, Xianzheng Ma, Xupeng Miao, Xuming He, and Bin Cui.
\newblock {CALIP}: Zero-shot enhancement of clip with parameter-free attention.
\newblock In \emph{AAAI}, 2023.

\bibitem[Jia et~al.(2021)Jia, Yang, Xia, Chen, Parekh, Pham, Le, Sung, Li, and Duerig]{ALIGN}
Chao Jia, Yinfei Yang, Ye Xia, Yi-Ting Chen, Zarana Parekh, Hieu Pham, Quoc Le, Yun-Hsuan Sung, Zhen Li, and Tom Duerig.
\newblock Scaling up visual and vision-language representation learning with noisy text supervision.
\newblock In \emph{ICML}, 2021.

\bibitem[Krause et~al.(2013)Krause, Stark, Deng, and Fei-Fei]{cars}
Jonathan Krause, Michael Stark, Jia Deng, and Li Fei-Fei.
\newblock 3d object representations for fine-grained categorization.
\newblock In \emph{ICCV workshops}, 2013.

\bibitem[Krizhevsky et~al.(2009)Krizhevsky, Hinton, et~al.]{cifar}
Alex Krizhevsky, Geoffrey Hinton, et~al.
\newblock Learning multiple layers of features from tiny images.
\newblock 2009.

\bibitem[Li et~al.(2023)Li, Prabhudesai, Duggal, Brown, and Pathak]{li2023your}
Alexander~C Li, Mihir Prabhudesai, Shivam Duggal, Ellis Brown, and Deepak Pathak.
\newblock Your diffusion model is secretly a zero-shot classifier.
\newblock In \emph{ICCV}, 2023.

\bibitem[Lin(2004)]{ROUGE}
Chin-Yew Lin.
\newblock {ROUGE}: A package for automatic evaluation of summaries.
\newblock In \emph{Workshop on Text Summarization Branches Out}, 2004.

\bibitem[Liu et~al.(2019)Liu, Ott, Goyal, Du, Joshi, Chen, Levy, Lewis, Zettlemoyer, and Stoyanov]{RoBERTa}
Yinhan Liu, Myle Ott, Naman Goyal, Jingfei Du, Mandar Joshi, Danqi Chen, Omer Levy, Mike Lewis, Luke Zettlemoyer, and Veselin Stoyanov.
\newblock {RoBERTa}: A robustly optimized bert pretraining approach.
\newblock \emph{arXiv preprint arXiv:1907.11692}, 2019.

\bibitem[Menon and Vondrick(2023)]{menon2022visual}
Sachit Menon and Carl Vondrick.
\newblock Visual classification via description from large language models.
\newblock \emph{ICLR}, 2023.

\bibitem[Parkhi et~al.(2012)Parkhi, Vedaldi, Zisserman, and Jawahar]{pets}
Omkar~M Parkhi, Andrea Vedaldi, Andrew Zisserman, and CV Jawahar.
\newblock Cats and dogs.
\newblock In \emph{CVPR}, 2012.

\bibitem[Popp et~al.(2024)Popp, Metzen, and Hein]{popp2024zero}
Niclas Popp, Jan~Hendrik Metzen, and Matthias Hein.
\newblock Zero-shot distillation for image encoders: How to make effective use of synthetic data.
\newblock \emph{arXiv preprint arXiv:2404.16637}, 2024.

\bibitem[Pratt et~al.(2023)Pratt, Covert, Liu, and Farhadi]{CuPL}
Sarah Pratt, Ian Covert, Rosanne Liu, and Ali Farhadi.
\newblock What does a platypus look like? generating customized prompts for zero-shot image classification.
\newblock In \emph{ICCV}, 2023.

\bibitem[Radford et~al.(2021)Radford, Kim, Hallacy, Ramesh, Goh, Agarwal, Sastry, Askell, Mishkin, Clark, et~al.]{CLIP}
Alec Radford, Jong~Wook Kim, Chris Hallacy, Aditya Ramesh, Gabriel Goh, Sandhini Agarwal, Girish Sastry, Amanda Askell, Pamela Mishkin, Jack Clark, et~al.
\newblock Learning transferable visual models from natural language supervision.
\newblock In \emph{ICML}, 2021.

\bibitem[Sanh et~al.(2019)Sanh, Debut, Chaumond, and Wolf]{DistilBERT}
Victor Sanh, Lysandre Debut, Julien Chaumond, and Thomas Wolf.
\newblock {DistilBERT}, a distilled version of {BERT}: smaller, faster, cheaper and lighter.
\newblock \emph{arXiv preprint arXiv:1910.01108}, 2019.

\bibitem[Touvron et~al.(2023)Touvron, Lavril, Izacard, Martinet, Lachaux, Lacroix, Rozi{\`e}re, Goyal, Hambro, Azhar, et~al.]{LLaMA}
Hugo Touvron, Thibaut Lavril, Gautier Izacard, Xavier Martinet, Marie-Anne Lachaux, Timoth{\'e}e Lacroix, Baptiste Rozi{\`e}re, Naman Goyal, Eric Hambro, Faisal Azhar, et~al.
\newblock {LLaMA}: Open and efficient foundation language models.
\newblock \emph{arXiv preprint arXiv:2302.13971}, 2023.

\bibitem[Van~der Maaten and Hinton(2008)]{van2008visualizing}
Laurens Van~der Maaten and Geoffrey Hinton.
\newblock Visualizing data using t-{SNE}.
\newblock \emph{Journal of Machine Learning Research}, 9\penalty0 (11), 2008.

\bibitem[Xiao et~al.(2010)Xiao, Hays, Ehinger, Oliva, and Torralba]{sun1}
Jianxiong Xiao, James Hays, Krista~A Ehinger, Aude Oliva, and Antonio Torralba.
\newblock Sun database: Large-scale scene recognition from abbey to zoo.
\newblock In \emph{CVPR}, 2010.

\bibitem[Xiao et~al.(2016)Xiao, Ehinger, Hays, Torralba, and Oliva]{sun2}
Jianxiong Xiao, Krista~A Ehinger, James Hays, Antonio Torralba, and Aude Oliva.
\newblock Sun database: Exploring a large collection of scene categories.
\newblock \emph{International Journal of Computer Vision}, 119:\penalty0 3--22, 2016.

\bibitem[Zhou et~al.(2017)Zhou, Lapedriza, Khosla, Oliva, and Torralba]{places}
Bolei Zhou, Agata Lapedriza, Aditya Khosla, Aude Oliva, and Antonio Torralba.
\newblock Places: A 10 million image database for scene recognition.
\newblock \emph{IEEE transactions on pattern analysis and machine intelligence}, 40\penalty0 (6):\penalty0 1452--1464, 2017.

\end{thebibliography}
}

\end{document}